\pdfoutput=1

\documentclass[11pt]{article}

\usepackage[]{acl}

\usepackage{times}
\usepackage{latexsym}

\usepackage[T1]{fontenc}

\usepackage[utf8]{inputenc}

\usepackage{microtype}
\usepackage{amsmath}
\usepackage{graphicx}
\usepackage{bbm}
\usepackage{enumitem}
\usepackage{dblfloatfix}
\usepackage{svg}

\title{Interpretation Quality Score for Measuring the Quality of Interpretability Methods}

\author{Sean Xie, Soroush Vosoughi, Saeed Hassanpour}

\begin{document}
\maketitle
\begin{abstract}
Machine learning (ML) models have been applied to a wide range of natural language processing (NLP) tasks in recent years. In addition to making accurate decisions, the necessity of understanding how models make their decisions has become apparent in many applications. To that end, many interpretability methods that help explain the decision processes of ML models have been developed. Yet, there exists no widely-accepted metric to evaluate the quality of explanations generated by these methods. As a result, there currently is no standard way of measuring to what degree an interpretability method achieves an intended objective. Moreover, there is no accepted standard of performance by which we can compare and rank the current existing interpretability methods. In this paper, we propose a novel metric for quantifying the quality of explanations generated by interpretability methods. We compute the metric on three NLP tasks using six interpretability methods and present our results. 
\end{abstract}

\section{Introduction}
Machine Learning models have recently achieved promising results in a wide variety of NLP tasks. \cite{khan2016survey} As a result of their performance, ML models have been utilized in a variety of domains, from health care \cite{ahmad2018interpretable} and drug discovery \cite{jimenez2020drug} to criminal justice \cite{rudin2019stop}, to aid human decision-making. Due to the underlying risks of each decision, the deployment of ML models are hinged on how well humans can understand the ML models' decision processes. Therefore, in addition to model performance, model interpretability has also become an area of interest for the ML research community. \cite{doshivelez2017rigorous}

\begin{figure}[!t]
   \includegraphics[]{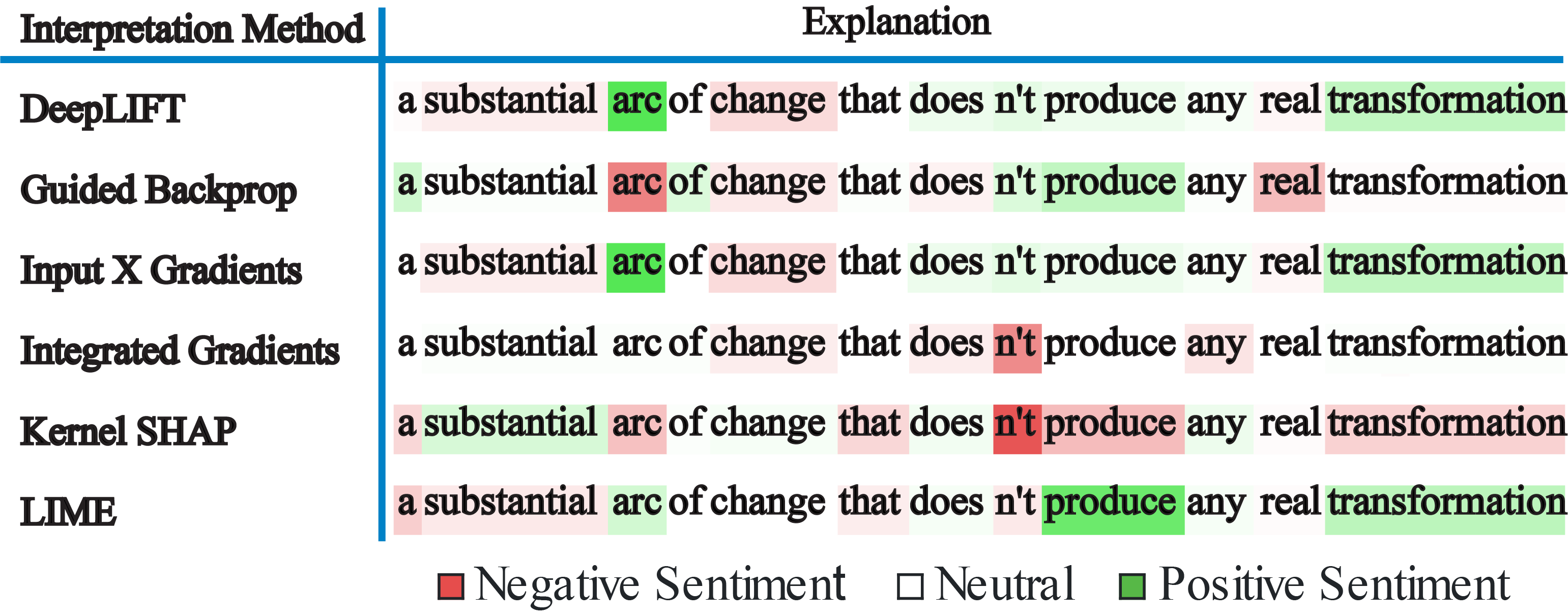}
    \caption{Six different interpretation methods yielding six different explanations for the same negative movie review.}
\end{figure}

Currently, an issue in the interpretability domain is that there is no established metric for measuring \emph{the quality of explanations} generated by the various existing interpretability methods. Without such a metric, it is difficult to benchmark and compare interpretability methods. This is problematic for ML practitioners who have an interest in selecting the best interpretability method for their applications. Moreover, the lack of a common, optimizable objective to work towards impedes progress of the entire research community in developing new interpretability methods and improving existing ones.

\begin{figure*}[!t]
\includegraphics[]{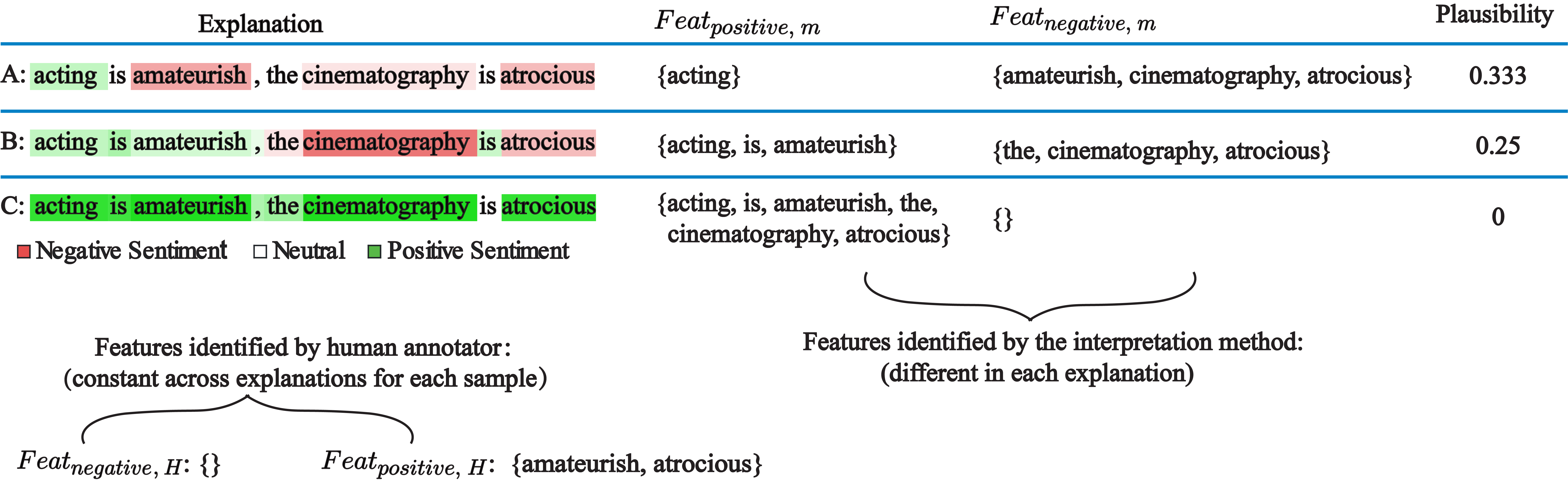}
\caption{Three example explanations generated by interpretability methods that have varying plausibility}
\end{figure*}

Let us examine the example in Figure 1. When the movie review ``a substantial arc of change that doesn't produce any real transformation'' is passed through a model, six different interpretability methods provide six different explanations for why the review is deemed negative by the model. A word like ``arc'' is deemed to have made a strong, positive contribution (i.e. carry a positive sentiment) to the output by DeepLIFT  \citep{shrikumar2016not}  and Input X Gradients \citep{DBLP:journals/corr/ShrikumarGK17}, while it is considered to have made a strong, negative contribution (i.e. carry a negative sentiment) by Guided Backprop \citep{springenberg2014striving}. LIME \citep{DBLP:journals/corr/RibeiroSG16} and Kernel SHAP \citep{DBLP:journals/corr/LundbergL17} identify ``arc'' as having positive and negative contributions, respectively, but the strength of their contributions are less than that of DeepLIFT and Input X Gradients, respectively. Finally, Integrated Gradients \citep{DBLP:journals/corr/SundararajanTY17} considers ``arc" to have made no impact on the model's decision at all, positively or negatively. In order to answer the question of ``which one of these explanations is the best?'', we first must have an establish metric of what is considered ``good'', i.e. a metric for the quality of explanations.

 The primary contributing factor to the lack of an interpretation quality metric is the lack of a general consensus on the definition and scope of interpretability itself. \cite{DBLP:journals/corr/Lipton16a} For example, \citet{slack2019assessing} considers interpretability to be the ability of a human to reproduce the model's results accurately given the input (simulatability). On the other hand, \citet{DBLP:journals/corr/abs-2004-03685} identifies
 how convincing the explanation is to humans (plausibility) to be the an important factor when evaluating interpretability. \citet{carvalho2019machine} and \citet{holland2020black} identify how well can humans understand the interpretation (comprehensibility) as attributes of interpretability. \citet{schmidt2019quantifying} attempts to define and quantify interpretability via measuring human trust and complexity of interpretations. At the moment, the word ``interpretability'' does not have an exact definition or a precise scope. Interpretability is instead a catch-all word that encompasses a plethora of properties and criteria that are coveted, for different reasons, in the pursuit of understanding of machine learning models \cite{DBLP:journals/corr/abs-1910-10045}. Therefore, when evaluating the explanations of different interpretability methods and comparing them , many current approaches are applied in an ad-hoc manner, with experiments designed to focus on the most sought-after attributes for that particular application \cite{herman2017promise}. 
 
In this work, we propose Interpretation Quality Score (IQS), a metric that quantifies the quality of interpretability methods on NLP tasks. IQS is a human and functionally-grounded composite metric that consists of 3 sub-metrics. Each of these 3 sub-metrics  measures the quality of the interpretability method under different criteria. We then evaluate the IQS of 6 popular interpretability methods  on 3 GLUE \citep{DBLP:journals/corr/abs-1804-07461} datasets (SST2, STSB and QNLI).

Our contributions in this study are as follows: \\
1. Introducing Interpretation Quality Score (IQS), a generalizable metric to quantify the quality of explanations generated by interpretability methods in for NLP. \\
2. Presenting an interpretability benchmark based on the IQS of 6 interpretability methods.

\begin{figure*}[!t]
 \centering
\includegraphics[]{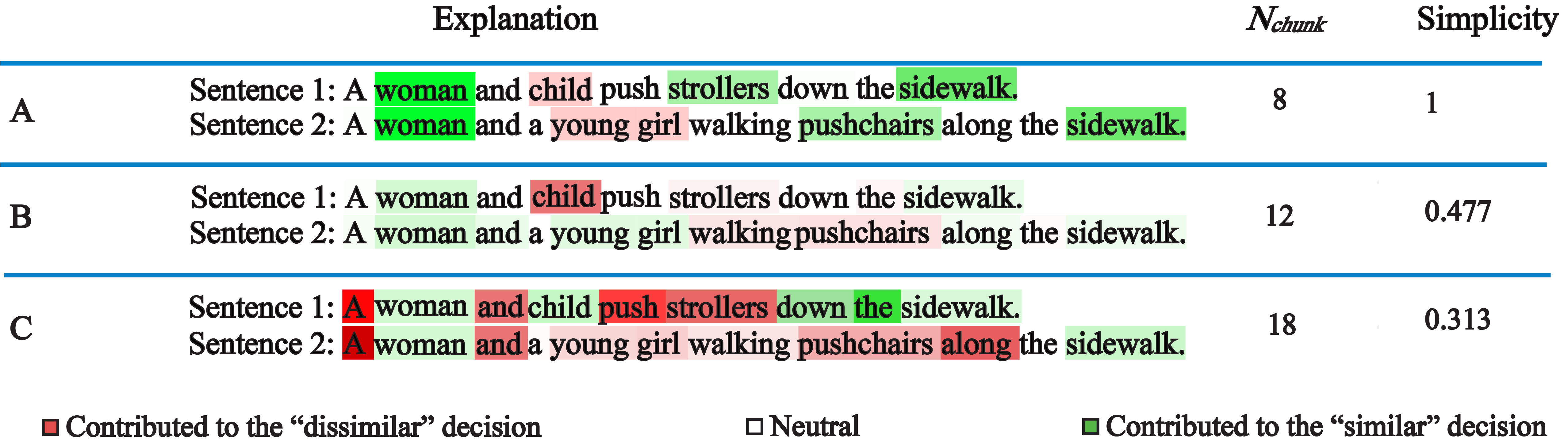}
\caption{An example from the Semantic Textual Similarity dataset, where each sample consists of two sentences and a score $\in [0, 5]$ of their semantic similarity. In this example, sentences 1 and 2 have a predicted similarity by the model of 4.3. Three explanations for the model's decision are provided. Words colored in green are words considered by the method to have caused the an increase in the predicted similarity score, i.e. made the model think the two sentences were similar. Words colored in red are words considered by the method to have caused a decrease in the predicted similarity score, i.e., made the model think the two sentences were dissimilar.}
\end{figure*}

\section{What makes an explanation good?}
In this work we contend that a "good" explanation that is provided by an interpretability method should be plausible, simple and make the underying ML model's decision reproducible for humans. IQS, therefore, evaluates the quality of an explanation based on three criteria: \textbf{plausibility}, \textbf{simplicity} and \textbf{reproducibility}. In this section we provide a description and an intuitive motivation for each of these three terms. The details on the calculation of each term are in section 3. 

\subsection{Plausibility}
The plausibility of an explanation is how sound and cogent the explanation  seems to a human. In Figure 2, the movie review, ``acting is amateurish, the cinematography is atrocious'', has been fed through and deemed negative by a model. Three different interpretability methods generate three different explanations for this decision. Explanation A marked the word "amateurish" and "atrocious" as having negative sentiments. This makes explanation A plausible because both "amateurish" and "atrocious" are being used to describe the review negatively. Explanation C, on the other hand, is not plausible because it has marked every word in the review as having positive sentiments, which is not convincing to the human annotator in this example.  Explanation B is less plausible than explanation A because it marks the word "amateurish" as having positive sentiments, but it is still more plausible than C because it at least still marks "atrocious" as having negative sentiments. 

\subsection{Simplicity}
Whether or not an explanation can be easily understood by a human is an important factor to consider when evaluating the quality of an explanation. It is said that human beings are only able to comprehend $7 \pm  2$ pieces of information at a time \cite{miller1956magical}. In this work, we utilize the complexity of an explanation as a proxy to its ability to be comprehended by humans. It is possible for interpretability methods to provide explanations that are too complex that they become unhepful. In Figure 3, explanation A marks the presence of ``woman'' and ``sidewalk'' in both sentences as making the model think that the two sentences are similar. Explanation A also marks ``child'' in sentence 1 and ``young girl'' in sentence 2 as two words that made the model think that the two sentences are dissimilar. Note that A also sheds light on the fact that ``strollers'' and ``pushchairs'' are considered to be semantically similar by the model. Explanation A highlights only 8 words and utilizes the least amount of information to explain the model's decision when compared explanations B and C; it therefore has the highest simplicity score out of the three.  Explanation B marks an additional 4 words (``the'', ``and'', ``walking'', ``along'') when compared to explanation A while explanation C marks nearly every word in the sentence pair as having some degree of contribution. Their complexities make them more difficult to understand than explanation A. 

\subsection{Reproducibility 
\footnote{Reproducibility is commonly referred to as "Simulatability" in interpretability literature \cite{doshivelez2017rigorous}.}
}
The purpose of interpretability methods is to explain the decision-making process of a ML model to a human. As a result, works such as  \citet{slack2019assessing} and \citet{doshivelez2017rigorous} have pointed to how well humans can correctly simulate the underlying model's decision when presented with an explanation and input as an important criterion when evaluating quality of explanations. However, reproducibility, when compared with plausibility and simplicity requires more human input to evaluate. Let us consider the movie review in Figure 2 and its accompanying three explanations. It is likely difficult - but not impossible - for someone who is presented with explanation C to classify the review as ``negative'', which is the model's actual decision. There is no way of collecting this information without running experiments involving human annotators. For the example in Figure 3,  how high of a similarity score would a human give when presented with explanations A, B, and C? The conduciveness of each explanation when it comes to helping humans reproduce the model's output is ambiguous is ambiguous in this case. As a metric, IQS therefore strongly relies on the help of human when evaluating reproducibility. 

\begin{table*}
\begin{align} 
\small
   \textup{IQS}                 = 
                                \alpha_{1} \cdot 
                                \underbrace{\frac{1}{|C|}\sum_{i \in C}
                                \frac{|Feat_{i, H} \cap Feat_{i, m}|}
                                     {|Feat_{i, H} \cup Feat_{i, m}|}}_
                                {\textup{Plausibility}}  
                                +\;
                                \alpha_{2} \cdot 
                                \underbrace{\frac{1}{\ln(N_{chunk}-\beta)+1}}_
                                {\textup{Simplicity}}  
                                +\;
                                \alpha_{3} \cdot 
                                \underbrace{\frac{1}{\mathcal{L}(y_H, y_M) + 1}}_ {\textup{ Reproducibility}}  
\label{eq:1}
\end{align}
\begin{center}

\end{center}
\end{table*}

\section{IQS}
We present the equation to IQS in equation 1. IQS $\in [0,1]$ is a composite metric that is the sum of three terms. Each of the three terms have output in the range $[0, 1]$ and is scaled by coefficients $\alpha_1, \alpha_2, \alpha_3$, respectively. These terms measure the quality of interpretability methods under separate criteria. The $\alpha$'s are in the interval $[0, 1]$ and have the additional property that $\alpha_1 + \alpha_2 + \alpha_3 = 1$.  The $\alpha$'s give the ML practitioner the ability to tune IQS calculation in such a way that is sensitive to the output of a particular task and reflective of the desired criterion for that task. 

\subsection{Calculating Plausibility} For each output class, $Feat_{i, H}$ stands for the set of features the human annotator identified to be indicative of the class (i.e, contributing to outputting that class as a decision) and $Feat_{i, m}$ stands for the set of features the interpretability method identified to be important for that class. For regression and generation tasks, there are two sets of $Feat_{i, H}$ and $Feat_{i, m}$,  representing features that are making positive contributions to the output and features that are making negative contributions to the model's output, respectively. The plausibility term calculates the mean of the Jaccard similarity coefficients between human-identified features and method-identified features across all classes. The plausibility term will be 1 when the interpretability method is able to identify exactly the set of features that the human annotator considers important for every class.

\subsection{Calculating Simplicity} Interpretability methods produce varying amount of information when explaining a model's decisions. We denote each unit of information the interpretability method produces as a  $N_{chunk}$, which is short for a cognitive chunk. In our NLP tasks, each  $N_{chunk}$ is a word that has been returned by the explanation as having some contribution. In general, an explanation's simplicity decreases when the amount of information needed to generate the explanation increases. For example, when the number of features and rules become too large, inherently interpretable models like Linear Regression and Decision Trees become incomprehensible to humans and therefore uninterpretable \cite{molnar2020interpretable}.  The simplicity term measures the quality of the interpretability methods in terms of complexity by penalizing interpretability methods that have large amounts of cognitive chunks and become incomprehensible. The term $\beta$ is a variable that can be tuned to adjust the rate at which simplicity decreases. In our experiments we set $\beta = 9$ because human beings are said to be able comprehend $7 \pm  2$ pieces of information at at a time, we subtract 9 from  $N_{chunk}$. \textbf{For explanations that have equal to or less than 9 $N_{chunk}$, we assign a simplicity score of 1.} Any explanation with $N_{chunk} <= 10$, in effect, will have a simplicity of 1. 

\subsection{ Calculating Reproducibility}  For each sample that the interpretability method is explaining, $y_H$ and $y_M$ stand for the human annotator's output with help from the interpretability method and the output of the ML model, respectively. $\mathcal{L}$ is an unspecified loss function that measures the difference between two sets of outputs. Similarly, $\mathcal{L}(y_H, y_M)$ measures how well the human annotator did, under the guidance of the interpretability method, in terms of producing the output of the ML model that is being explained. The reproducibility term will be 1 when the human annotator is able to produce outputs that are both completely accurate and identical to that of the ML model's, i.e., when $\mathcal{L}(y_H, y_M) = 0$.  

\section{Related Work}
Current approaches for evaluating interpretability methods can be divided into three categories. \cite{doshivelez2017rigorous}.
\begin{itemize}[noitemsep]
    \item  \textbf{Application-grounded evaluation} requires conducting experiments with end-users on real-world applications. The experiments would directly test how well the objective of a certain task is achieved and this performance is indicative of the level of success of the explanation.  A limiting factor of this approach is that evaluation schemes and evaluation metrics designed based on a specific application are only useful for that application. 
    
     \item \textbf{Human-grounded evaluation} refer to conducting simpler human–subject experiments that maintain the essence of the target application \cite{zhou2021evaluating}. In contrast to application-grounded evaluation, human-grounded evaluation do not require the involved humans to be end-users. Lay people who do not possess either machine learning knowledge or domain knowledge can carry out the evaluations and assess the quality of explanations. This evaluation approach assesses only the quality of the explanations, and not the accuracy of the associated prediction. The success of human-grounded evaluations is considered to be highly-indicative of the success of the interpretability method. 
    
    \item \textbf{Functionality-grounded evaluation} is a general approach to evaluating interpretation quality that does not involve human experiments. Functionally-grounded evaluations rely on predetermined definitions of interpretability that will serve as a proxy for the quality of explanations. e.g., the depth of a decision tree. Functionally-grounded evaluation approaches are advantageous in that they usually do not requiring human input. 
   
\end{itemize}

For human-based experiments (application-grounded and functionally-grounded evaluations), explanation qualities are measured by both quantitative and qualitative metrics. Quantitative metrics involve measuring the performance (e.g. accuracy) of human-produced results based on the output of intepretation methods. Qualitative metrics involve measuring  the satisfaction, confidence, and trust in interpretability methods via interviews and questionnaires \cite{poursabzi2021manipulating, zhou2016correlation, zhou2019effects}. 

For functionally-grounded evaluations, methods can be further divided into three types based on the underlying ML system of the interpretability method \cite{markus2021role}. 

\begin{itemize}[noitemsep]
    \item \textbf{Model-based explanations} are explanation methods that use simpler, surrogate models to explain the more complex original model. Evaluation metrics for these methods depend on the original model and surrogate models involved, but it is common for them to involve measures of complexity and simulatability. (e.g., number of rules for a decision tree, depth of the tree, number of Boolean operations needed for a human to run the model) \cite{guidotti2018survey, molnar2020interpretable, slack2019assessing}.
    
    \textbf{Attribution-based explanations} attributes influences of the model's predictions to input features. Many approaches for evaluating these interpretability methods involve calculating the degree of similarity between features recovered by the method and a set of ground truth features. These methods also use the recovered features to calculate proxy values for simulatability and fidelity \cite{DBLP:journals/corr/RibeiroSG16, ancona2017towards, sundararajan2017axiomatic}. 
    
    \item \textbf{Example-based explanations} select instances from the training/testing dataset or create new instances to create summarized explanations of a model. \citet{nguyen2020quantitative} identified representativeness of explanations and diversity of samples as measures for fidelity and the degree of integration of the explanation. In addition, the simplicity of the explanation (the number of examples used) is a proxy  for the ease with which humans can process the explanation. 
    
\end{itemize}

\subsection{Where does our metric stand?}
IQS is an interpretability metric based on both functionally-grounded evaluations and human-grounded evaluations.  IQS requires obtaining human annotations for interpretations. The annotations are then used in a functionally-grounded approach to calculate the degree to which interpretability methods satisfy certain axioms and properties. IQS is meant to be generalizable and applicable across NLP tasks and potential ML applications. Therefore, we do not employ application-grounded evaluation approaches in IQS. However, IQS has elements that are inspired by application-grounded evaluation in the form of hyperparameters that one can tune in order to better capture and reflect the sub-metrics within IQS that are more desired for a specific application. 

\section{Evaluated Interpretability Methods}
In this section we provide brief overviews of interpretability methods we evaluated and benchmarked using IQS.  We note here that, for our NLP tasks, the features identified by the interpretability methods listed below as well the features identified by human annotators are both words. Our implementations of these interpretability methods are based on the PyTorch library Captum \cite{kokhlikyan2020captum}. 

Input X Gradient is a simple feature attribution approach that extends the notion of saliency. Input X Gradient works by multiplying each feature in the input with their respective gradient \cite{shrikumar2016not}.  

Guided Backpropagation computes the gradient of the target's output with respect to the input. However, ReLU activations' gradients are overriden and only zero and positive gradients are propagated back \cite{springenberg2014striving}.

LIME draws samples of data by making minor perturbations around the input and trains an interpretable surrogate model \cite{DBLP:journals/corr/RibeiroSG16}. For our experiments, we set the number of points to sample to 3000 and use a Lasso Regression with regularizing strength 3e-4 as our interpretable surrogate model. 

Kernel SHAP uses the LIME framework to compute Shapley Values. The correct configuration of loss function, weighting kernel and regularization terms in the LIME framework allow for more efficient calculation of Shapley Values \cite{DBLP:journals/corr/LundbergL17}. We use the same LIME configuration for Kernel SHAP. 

DeepLIFT is a back-propagation based method. It calculates and attributes variations to the input using the differences between the inputs and corresponding references/baselines for non-linear activations \cite{DBLP:journals/corr/ShrikumarGK17}. For our experiments, we use a zero scalar as our baseline. 

Integrated Gradients assigns importance scores to input features via computing integral approximations of gradients with respects to inputs along the path from a provided reference/baseline \cite{DBLP:journals/corr/SundararajanTY17}. For our experiments, we set the number of steps for the approximation method to 50 and use a zero scalar as our baseline. 

\begin{figure*}
\includegraphics[]{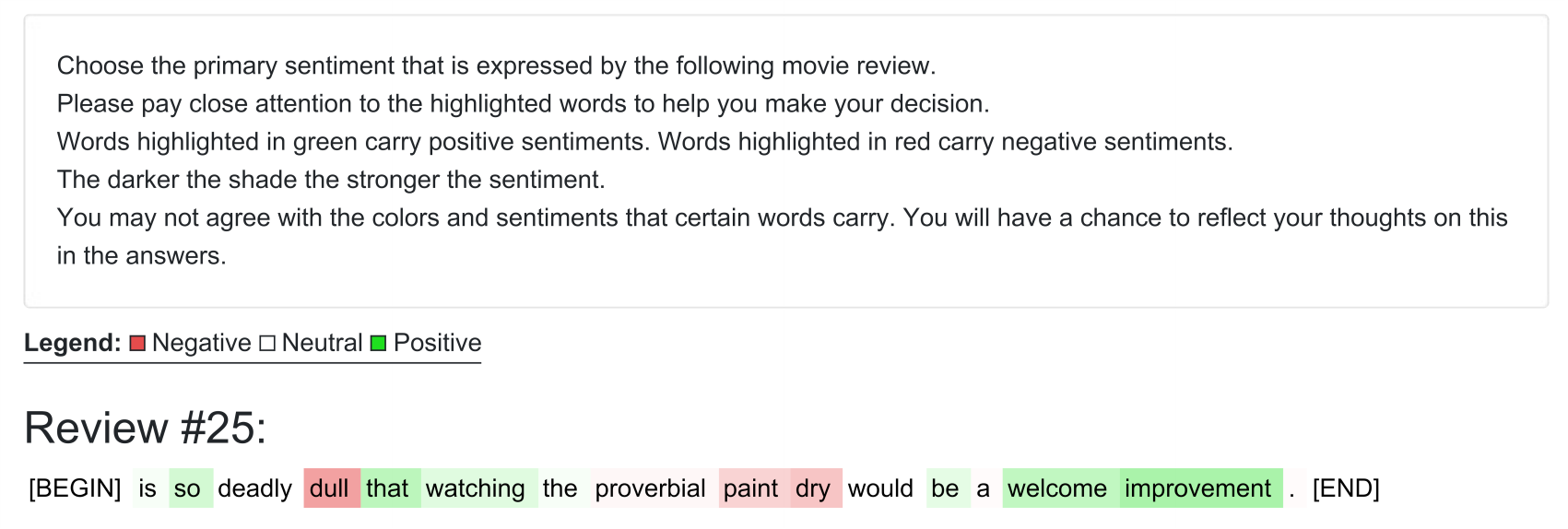}
\caption{MTurk instruction for a SST2 example and corresponding explanation generated by Input X Gradient.}
\end{figure*}

\section{Experiments}

We evaluated the quality of the 6 interpretability methods on the SST2, STSB and QNLI datasets from GLUE, an English NLP benchmark, to perform the following 3 NLP tasks: Sentiment Analysis,  Semantic Textual Similarity, and Question Natural Language Inference. We hired workers from Amazon Mechanical Turk (MTurk) to be our human annotators. For all tasks, the underlying language model the interpretability methods are trying to explain was RoBERTa \cite{DBLP:journals/corr/abs-1907-11692}. For each task, dense layers with appropriate activation functions are attached and trained on top of RoBERTa to perform regression for that task. Regression outputs are then used for classification purposes based on a threshold value. $\mathcal{L}$ in reproducibility for SST2 and QNLI was log loss and $\mathcal{L}$ for STSB was mean absolute error.

For each task, we ran the model over 50 randomly drawn samples (25 from each of the two classes) and applied the 6 explanation methods to explain model outputs. We obtained $y_M$ in the reproducibility term from the outputs of the model.  Then, for each interpretability method, we obtained $Feat_{i, m}, \; \forall i \in C$ in the plausibility term as well as $N_{chunk}$ in the simplicity term from the explanations generated by the interpretability method. We note here that, since the input features to the model are individual words, the interpretability methods attribute importance to each individual word. We therefore considered each individual word that has been identified by the interpretability method to be a cognitive chunk.

In order to obtain $y_H$ in the reproducibility term and $Feat_{i, H},\; \forall i \in C$ in the plausibility term, we first created MTurk tasks using the samples and their corresponding explanations. We then asked our annotators to annotate the samples with the help of the explanations. We also asked our annotators to identify words they think that should be indicative of each class. An example explanation generated by Input X Gradient on a sample from SST2 and its corresponding MTurk layout is provided in Figure 1. After showing the annotator the information depicted in Figure 1, we asked the annotator the following questions. 

\begin{itemize}[noitemsep]
    \item Question 1) The primary sentiment expressed by the review was? (Negative/Positive) 
    \item Question 2) In the review, are there any words that are highlighted in RED that you think should NOT be highlighted in RED? (Are there RED words that you do not consider to be negative words?) 
    \item Question 3) In the review, are there any words that are highlighted in GREEN that you think should NOT be highlighted in GREEN? (Are there GREEN words that you do not consider to be positive words?) 
    \item Question 4) In the review, are there any words that are NOT highlighted in RED that you think should have been highlighted in RED? (Are there negative words that were not identified?) 
    \item Question 5) In the review, are there any words that are NOT highlighted in GREEN that you think should should have been highlighted in GREEN? (Are there positive words that were not identified?)
\end{itemize}
The questions' variations across the three tasks were minimal. For example, for QNLI, Question 1) asks whether or not there is entailment between the paragraph and the question posed instead of whether the sentiment was positive or negative. We assigned 3 annotators per sample and calculate the mean across the 3 annotators to obtain values for The reproducibility term and the plausibility term. We observed 63\% agreement rate across all samples between our annotators for SST2, 83\% agreement rate for STSB, and 65\% agreement rate for QNLI. 

\section{Results}
In Table 1, we report, for each task, the IQS of the 6 interpretability methods under the combination of $\alpha$'s such that Terms 1, 2 and 3 are given equal weight, i.e., $\alpha_1 = \alpha_2 = \alpha_3 = 1/3$. We note here that the ranking of the interpretability methods based on IQS is more indicative of interpretability than the magnitude of the IQS itself when comparing interpretability methods. 

In Table 2, we generate 66 unique combinations of  $\alpha_1, \alpha_2$ and $\alpha_3$ with values from the set $\{{0}\} \cup \{{x/10 \;|\; x \in \mathbbm{N},  x \leq 10}$\} and under the constraint that $\alpha_1 + \alpha_2 + \alpha_3 = 1$. (e.g. $\alpha_1 = 0.3, \alpha_2 = 0.4, \alpha_3 = 0.3$) and report the mean and standard deviation of IQS for each interpretability method across all 66 combinations for each task. We then use the average across all three tasks and create Figure, where we show the relative strength of each interpret ability method with regard to each criterion of IQS.


\begin{table*}
\centering
\begin{tabular}{l|c|c|c|c}
\hline
\textbf{Interpretability method} & \textbf{Plausibility} & \textbf{Simplicity} & \textbf{Reproducibility} & \textbf{IQS}\\
\hline
\hline
\multicolumn{5}{c}{Results for SST2}\\
\hline
\hline

\textbf{Input X Gradient}  & 0.2462 & 0.2466 & 0.2599 & 0.7527 \\ 
\textbf{DeepLIFT}  & 0.2430 & 0.2466 & 0.2621 & 0.7517 \\ 
\textbf{Kernel SHAP}  & 0.2555 & 0.2392 & 0.2556 & 0.7503 \\ 
\textbf{LIME}  & 0.1726 & 0.1920 & 0.2750 & 0.6397 \\ 
\textbf{Guided Back Propagation}  & 0.1784 & 0.1765 & 0.2664 & 0.6213 \\ 
\textbf{Integrated Gradients}  & 0.1698 & 0.1977 & 0.2599 & 0.6274 \\ 

\hline
\hline
\multicolumn{5}{c}{Results for STSB}\\
\hline
\hline

\textbf{Input X Gradient}  & 0.2437 & 0.1597 & 0.3089 & 0.7122 \\ 
\textbf{DeepLIFT}  & 0.2455 & 0.1597 & 0.3056 & 0.7107 \\ 
\textbf{Kernel SHAP}  & 0.2392 & 0.1192 & 0.3056 & 0.6639 \\ 
\textbf{Guided Back Propagation}  & 0.2232 & 0.1137 & 0.3133 & 0.6502 \\ 
\textbf{LIME}  & 0.2108 & 0.1158 & 0.3189 & 0.6454 \\ 
\textbf{Integrated Gradients}  & 0.1306 & 0.0971 & 0.3133 & 0.5410 \\ 

\hline
\hline
\multicolumn{5}{c}{Results for QNLI} \\
\hline
\hline
\textbf{Kernel SHAP}   & 0.2848 & 0.0945 & 0.2405 & 0.6198 \\ 
\textbf{Input X Gradient}   & 0.3052 & 0.0885 & 0.2232 & 0.6169 \\ 
\textbf{Guided Back Propagation}   & 0.2888 & 0.0854 & 0.2426 & 0.6168 \\ 
\textbf{DeepLIFT} & 0.2863 & 0.0885 & 0.2405 & 0.6153 \\ 
\textbf{Integrated Gradients}  & 0.2644 & 0.1075 & 0.2232 & 0.5950 \\ 
\textbf{LIME}  & 0.2834 & 0.0808 & 0.2275 & 0.5917 \\ 

\hline
\hline
\end{tabular}
\caption{Scaled The reproducibility term, the plausibility term, and the simplicity term and the corresponding IQS when  $\alpha_1 = \alpha_2 = \alpha_3 = 1/3$}
\label{tab:accents}
\end{table*}

\begin{figure*}[!b]
 \centering
 \includegraphics[]{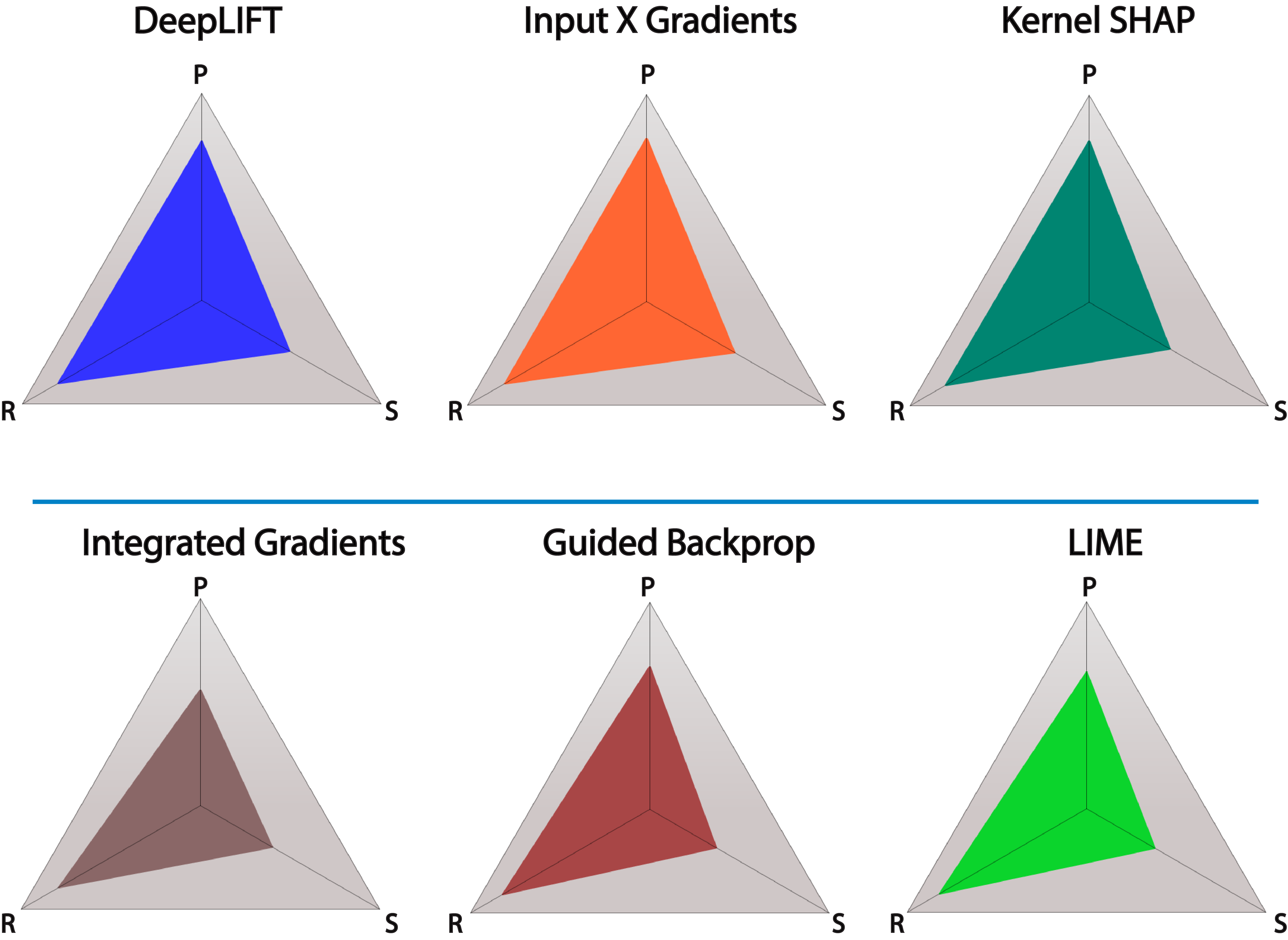}
    \caption{Average score of each interpretability method across all tasks for each criterion. P for Plausibility, R for Reproducibility and S for Simplicity. }
\end{figure*}

\begin{table*}[!t]
\centering
\begin{tabular}{l|c|c|c}
\hline
\textbf{Interpretability method} & \multicolumn{3}{c}{\textbf{IQS Mean \small (std)}} \\ 
\hline
\textbf{} & \textbf{SST2} & \textbf{STSB} & \textbf{QNLI} \\ \cline{2-4} 
\textbf{Input X Gradient}& 0.7527  \small (0.0191) & 0.7122 \small(0.1832) & 0.6169  \small(0.2680) \\ 
\textbf{DeepLIFT} & 0.7517  \small (0.0248) & 0.7107 \small(0.1796) & 0.6153 \small(0.2537) \\ 
\textbf{Kernel SHAP} & 0.7503 \small (0.0231) & 0.6639  \small(0.2314) & 0.6198  \small(0.2439) \\

\textbf{LIME} & 0.6397  \small (0.1332) & 0.6454 \small (0.2489) & 0.5917 \small(0.2564) \\

\textbf{Integrated Gradients} & 0.6274  \small (0.1130) & 0.5410  \small(0.2851) & 0.5950  \small(0.1992) \\

\textbf{Guided Back Propagation} & 0.6213   \small (0.1258) & 0.6502   \small(0.2449) & 0.6168   \small(0.2612)\\

\end{tabular}
\caption{Mean and Standard Deviation of IQS across all combinations of $\alpha$'s }
\label{tab:accents}
\end{table*}

\section{Discussion}
Interpretability in ML has garnered much attention recently. \cite{linardatos2021explainable} Many interpretability methods have been developed by the research community. However, without an established metric, obtaining objective measures of interpretability has proved to be difficult. Currently, benchmarks for existing interpretability methods are scarce. \citep{DBLP:journals/corr/abs-1801-05075} utilized a completely functionally-grounded approach to evaluate the quality of several gradient-based interpretability methods for Computer Vision tasks. \citep{DBLP:journals/corr/abs-1806-10758} proposed an approach to benchmark interpretability methods based on human-attention masks but only evaluated their approach on two interpretability methods. Although both works introduced benchmarking approaches generalizable outside of their applications, neither introduced a metric to quantify the quality of interpretations.  To our knowledge, our work is the first effort to introduce an interpretability metric and a benchmark on exisitng interpretability methods for tasks in the NLP domain. With an established and optimizable metric of interpretability, we hope to inspire more work in the direction of refining interpretability methods. 

IQS provides an universal way to rate and rank interpretability methods. Its terms and hyperparameters allow IQS to be tailored to various tasks that seek different criteria for model interpretability. Researchers and ML practitioners can utilize IQS to evaluate interpretability methods for their specific tasks, provided they adjust the hyperparameters appropriately.  In addition, they can choose to use IQS as a frame of reference when deciding what interpretability methods to employ for their task. We concede that there is no single metric that can cover every important aspect of interpretability. \cite{zhou2021evaluating} However, we believe that the aspects of interpretability covered by IQS are expansive enough such that IQS is generalizable and applicable to many domains and tasks.

The interpretability methods we evaluated in our work have algorithms with different levels of complexity. Notably, the intuition and algorithms behind Input X Gradient and Guided Back Propagation are simpler than the rest. Our experiments show that the complexity of an interpretability method is not reprentative of its IQS, as evident from the fact that Input X Gradient was consistently one of the two interpretability methods with the highest IQS.  \citet{holland2020black} states that the human capacity for understanding determines whether or not an AI system is interpretable. \citet{doshivelez2017rigorous} states that human evaluation is essential to assessing interpretability.  Interpretability is ultimately a human-based concept.

Confidence and trust are important factors in the adoption and utilization of ML models in high-stake applications. Much like measuring interpretability, quantifying confidence and trust directly is difficult. Our metric offers the ability to gauge how easy/difficult it is for humans to understand the ML model's decision making process. We hope our metric can inspire people to have increased confidence and trust in ML models. In  addition, we hope interpretability research based on our metric can bring about wider and safer adoption of ML models in many high-stake domains.

\section{Limitations and Future Work}
The interpretability methods we chose to benchmark are all feature-attribution methods. We chose them because they are popular, effective and produce many features (words) with which we can calculate the plausibility term and the simplicity term. We note here, however, that IQS is capable of being applied to interpretability methods that do not rely on feature attribution. For interpretability methods that do not return features as outputs, one may need to change their configuration of $\alpha$'s such that they don't place as much emphasis on the plausibility term and the simplicity term.

Alternatively, one may redefine what a ``feature'' or a ``cognitive chunk'' is for their particular application. For IQS, Features and cognitive chunks do not necessarily have to be part of the input vector. Features are what the interpretability method returns to highlight the decision-making process of the underlying ML model. A cognitive chunk, on the other hand, is the unit of information through which features are understood by humans. It happens so that in our application, both features and cognitive chunks were words. For a interpretability method that is surrogate-based and utilizes decision trees and lists, a feature could very well be a rule and a cognitive chunk be a single boolean operation in a rule.

The interpretability methods we evaluated have the potential to achieve higher IQS under different configurations of hyperparameters. For example, the selection of  baselines/references for DeepLIFT, regularization strength for LIME, and number of steps for the approximation algorithm for Integrated Gradients can all be finetuned. In this work, we performed a sensitivity analysis and used hyperparameters that are sensible for each interpretability method to obtain our results. \cite{kokhlikyan2020captum}. As the purpose of this work is to establish a metric, we leave fine-tuning intepretability methods in the pursuit of optimizing our metric to future work. 

\section{Conclusion}
In conclusion, establishing a metric to quantify the quality of interpretability methods is important for the widespread utilization of machine learning models. Such a metric would provide a common objective for researchers to optimize. Moreover, it would allow the benchmarking and comparison of existing interpretability methods. Finally, a metric to quantify the quality of interpretability methods can be used to gauge how well explanations and models are understood by humans, which in turn boosts people's confidence and trust in ML models. In this work, we introduced Interpretation Quality Score (IQS), a human- and functionally-grounded approach to measure the quality of explanations generated by interpretability methods. We evaluated the IQS of 6 interpretability methods (Input X Gradient, DeepLIFT, Kernel SHAP, Integrated Gradients, LIME and Guided Back Propagation ) on 3 NLP tasks (SST2, STSB, QNLI)  and present the benchmarks and rankings observed. 

\section{Experimental Details}
We used pre-trained RoBERTa models for each task from Hugging Face's transformers library.\footnote{Paper and repositories for pretrained models: \newline
\cite{heitmann2020} \newline
https://huggingface.co/cross-encoder/STSB-roberta-large \newline
https://huggingface.co/cross-encoder/QNLI-distilroberta-base
}
We ran our models and the interpretability methods on a Nvidia Titan RTX GPU with 24 GB of memory. The combined time to generate all sample outputs and interpretations was around 26 hours. We obtained a total of 2700 responses from our MTurk annotators, with each response answering all of questions 1-5.

\section{Ethics Statement}
For our experiments, we hired workers from Amazon Mechanical Turks. We restrict the location of workers to be within the United States (US) and allowed only master workers with Human Intelligence Task (HIT) approval rate greater than or equal to 99 to complete our tasks. We paid \$1.00 per task. Since the average completion time of a task was about 4 minutes, this translates to roughly around \$15 per hour,  which is significantly higher than the US federal minimum wage. Annotators were anonymous and not asked about their personal information.

\bibliographystyle{acl_natbib}
\bibliography{custom}

\appendix
\end{document}